\def\year{2020}\relax
\documentclass[letterpaper]{article} 
\usepackage{aaai20}  
\usepackage{times}  
\usepackage{helvet} 
\usepackage{courier}  
\usepackage[hyphens]{url}  
\usepackage{graphicx} 
\usepackage{amsmath}
\usepackage{graphicx}  
\title{Weakly-supervised Fine-grained Event Recognition on Social Media Texts for \\Disaster Management}

\author{Wenlin Yao\textsuperscript{1}, Cheng Zhang\textsuperscript{2}, Shiva Saravanan\thanks{Work was done while Shiva Saravanan was a local high school student and an intern in the NLP lab at Texas A\&M University.}\textsuperscript{3} 
{\bf \Large Ruihong Huang\textsuperscript{1}, Ali Mostafavi\textsuperscript{2}} \\ \textsuperscript{1}Department of Computer Science and Engineering, Texas A\&M University\\
\textsuperscript{2}Department of Civil Engineering, Texas A\&M University
\textsuperscript{3}Department of Computer Science, Princeton University\\
\{wenlinyao, czhang\}@tamu.edu, shivas@princeton.edu, huangrh@cse.tamu.edu,
amostafavi@civil.tamu.edu\\}

\begin{document}

\maketitle

\begin{abstract}
People increasingly use social media 
to report emergencies, 
seek help or share information during disasters, which makes social networks an important tool 
for disaster management.
To meet these time-critical needs, 
we present a weakly supervised approach for rapidly building high-quality classifiers 
that label each individual Twitter message with fine-grained event categories. 
Most importantly, we propose a novel method to create high-quality labeled data in a timely manner that  
automatically clusters tweets containing an event keyword 
and asks a domain expert to disambiguate event word senses and label  clusters quickly. 
In addition, to 
process extremely noisy and often rather short user-generated messages, we enrich tweet representations using preceding context tweets 
and reply tweets 
in building event recognition classifiers. 
The evaluation on two hurricanes, {\it Harvey} and {\it Florence}, shows that using only 1-2 person-hours of human supervision, the rapidly trained weakly supervised classifiers outperform supervised classifiers trained using more than ten thousand annotated tweets created in over 50 person-hours. 
\end{abstract}

\section{Introduction}

Due to its 
convenience, 
people increasingly use social media to report emergencies, provide real-time situation updates, offer or seek help or share information during disasters. 
During the devastating hurricane {\it Harvey} for example, the local authorities and disaster responders 
as well as the general public 
had frequently employed Twitter 
for real-time event sensing, 
facilitating evacuation operations, or finding victims in need of help. 
Considering the large volume of social media messages, it is necessary to achieve automatic recognition of life-threatening events based on individual messages 
for improving the use of social media during disasters.  This task is 
arguably more challenging than the well-studied 
collection-based event detection task on social media that often relies on detecting a burst of words over a collection of messages, especially considering the unique challenges of social media texts being extremely noisy and short. 


\begin{figure}[t]
 \begin{center}
 \includegraphics[width = 2.6in]{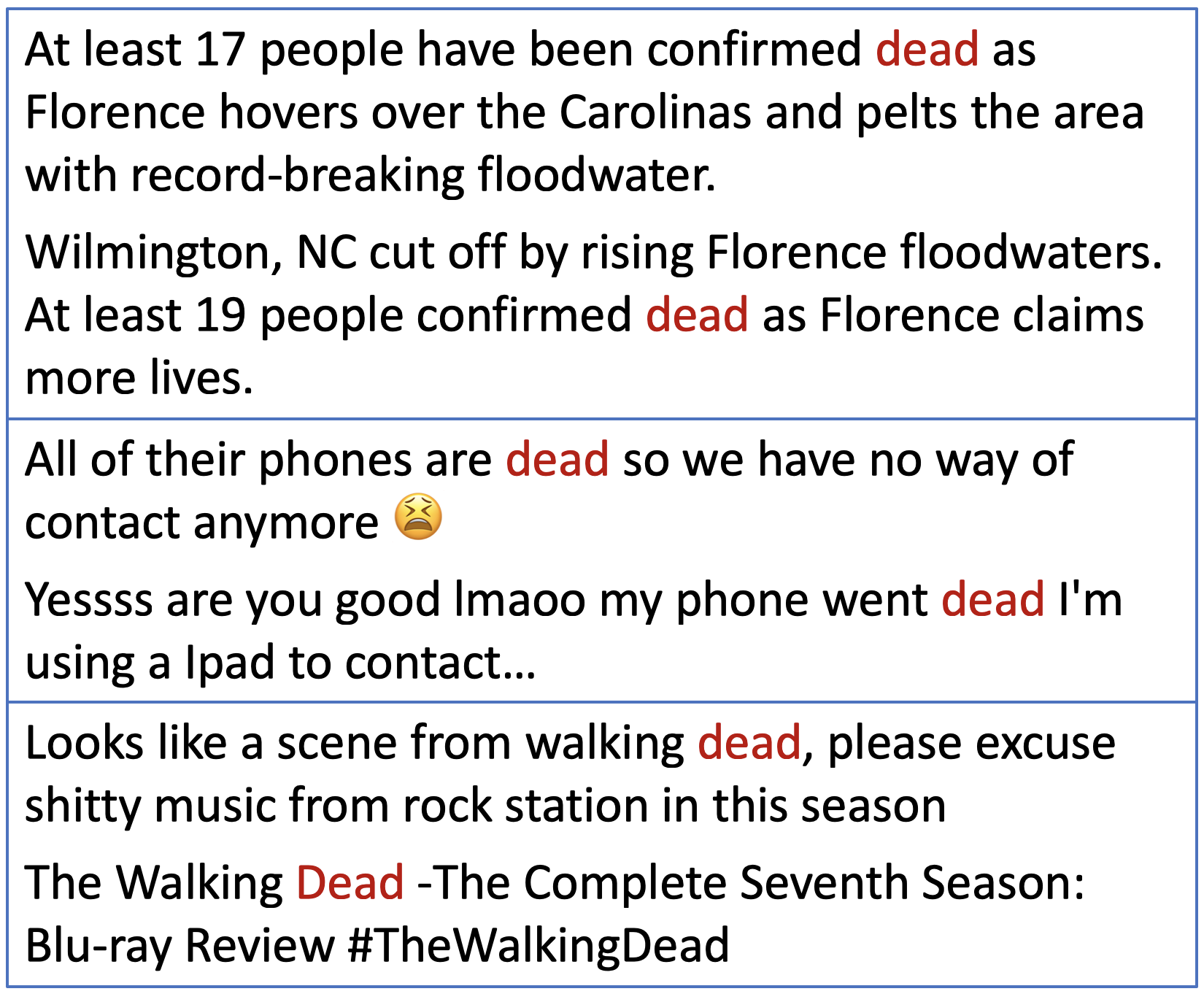}
  \end{center}
  \setlength{\belowcaptionskip}{-10pt}
 \caption{Examples of three senses of the word ``dead''}
\label{tweet_examples}
\end{figure}

To facilitate disaster management, especially during the time-critical disaster {\it response} phase, it is vital to build  event recognizers rapidly. 
However, the typical supervised learning paradigm 
requires a carefully labeled dataset 
that is normally created by asking human annotators to go through a large number of data instances and label them one by one, and the data labeling procedure usually takes days at least.
For fast deployment, 
we propose a novel data labeling method and an overall weakly supervised learning approach that quickly builds reliable fine-grained event recognizers. 

Specifically, to quickly label data, we explore the idea of 
identifying several high-quality event keywords and populating the keywords in a large unlabeled tweet collection. 
But, we quickly realize that it is essentially impossible to find an event keyword that is not ambiguous and has only one meaning in social media. 
Taking the word ``dead'' for example, in addition to the meaning of ``losing life'', ``dead'' is also frequently used to refer to phones being out of power or a TV series ``walking dead'', with example tweets shown in Figure \ref{tweet_examples}. 
It is a challenging problem 
because current automatic word sense disambiguation systems only achieve mediocre performance and may not work well on tweets with little in-domain training data, especially considering that many word senses appearing in tweets may even not appear in conventional sense inventories at all,  e.g., the word ``dead'' referring to the TV series ``walking dead''. 

Luckily, we observe that tweets adopting one common sense of an event keyword often share content words and can be easily grouped together. 
This observation is consistent with previous research on unsupervised word sense disambiguation \cite{yarowsky1995unsupervised,navigli2010experimental}. 
Therefore, we first 
cluster keyword identified noisy tweets using an automatic clustering algorithm and rank tweet clusters based on the number of tweets in each cluster. Next, we conduct manual Word Sense Disambiguation (WSD) by simply asking a domain expert to quickly go through the top-ranked clusters and judge whether each tweet cluster show the pertinent meaning of an event keyword, based on an inspection of five example tweets randomly sampled from a cluster. The domain expert is instructed to stop once 20 pertinent clusters have been identified. 
In this way, we significantly improved the quality of 
keyword identified tweets, requiring only 1-2 person-hours of manual cluster inspection time. 
Note that this is the only step in 
the overall weakly supervised approach that requires human supervision. 

\begin{figure}[t]
 \begin{center}
 \includegraphics[width = 3.0in]{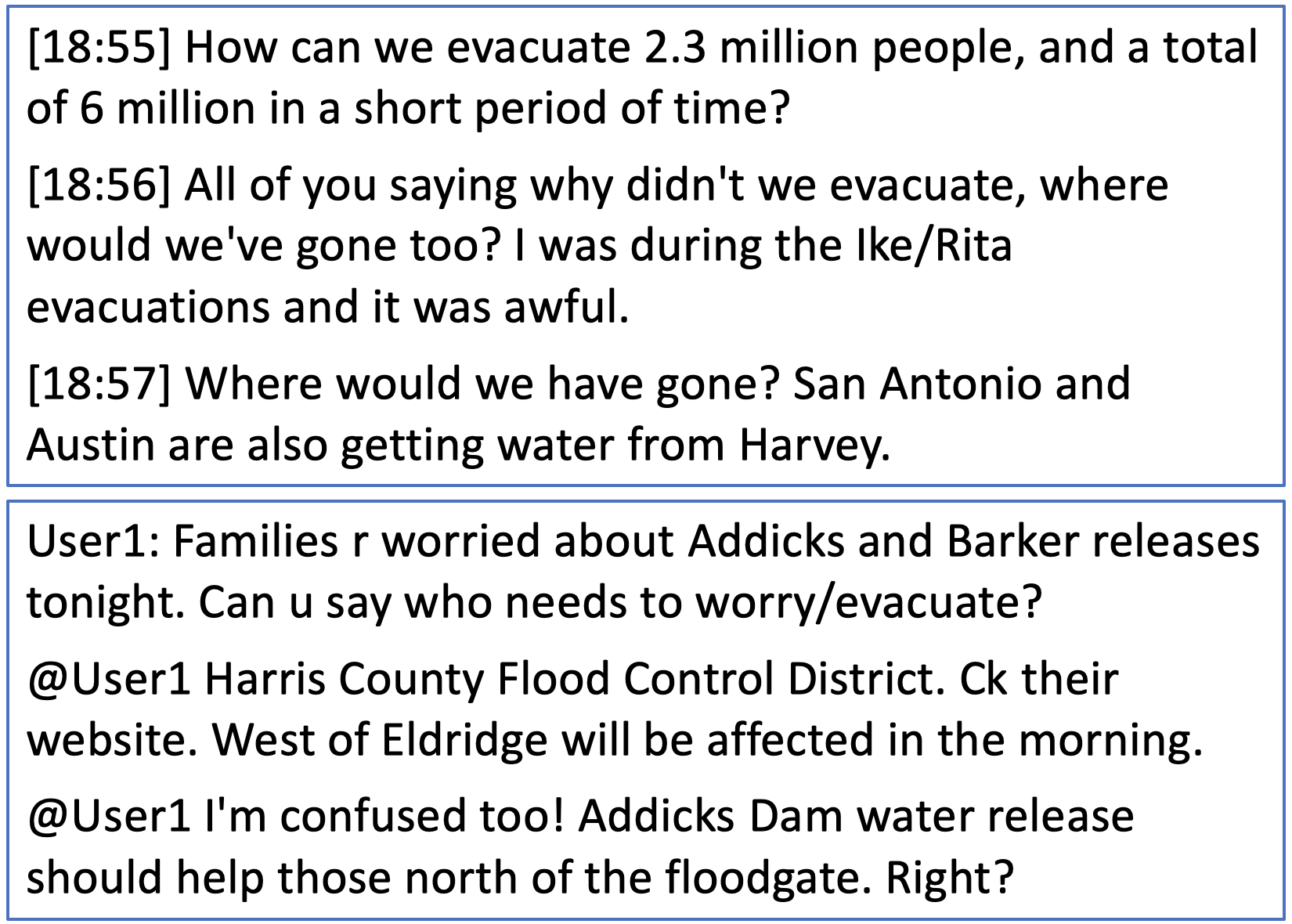}
  \end{center}
  \setlength{\belowcaptionskip}{-10pt}
 \caption{Examples with context and reply tweets}
\label{context_reply}
\end{figure}

Next, we use the rapidly created labeled data to train a recurrent neural net classifier and learn to recognize fine-grained event categories for individual Twitter messages. 
But tweets are often rather short, and it is difficult to make event predictions solely based on the content of a tweet itself. 
Instead, we use preceding context tweets posted by the same user as well as replies from other users, together with the target tweet,  
in a multi-channel neural net to predict the right event category.  
The observation is that the context tweets as well as reply tweets can both provide essential clues for inferring the topic of the target tweet. For instance, 
the upper example of Figure \ref{context_reply} shows that 
the two preceding tweets from the same user indicate the third tweet is asking about the location for evacuation; and the lower example shows that based on the reply tweet messages, we can infer  the first tweet is regarding water release of reservoir even having no external knowledge about Addicks/Barker. 


Finally, we further improve the multi-channel neural net classifier 
by applying it to label tweets and using the newly labeled tweets to augment the training set and retrain the classifier. The whole process goes for several iterations.  
The evaluation on two hurricane datasets, hurricane {\it Harvey} and {\it Florence}, shows 
that the rapidly trained weakly supervised systems\footnote{Our system is available at \url{https://github.com/wenlinyao/AAAI20-EventRecognitionForDisaster}.} using the novel data labeling method outperforms the supervised learning approach requiring thousands of carefully annotated tweets created in over 50 person-hours.

\section{Related Work}

Previous research for Twitter event detection mostly focuses on unsupervised statistical approaches, which can be categorized into three main streams. 
1) Identifying burst topics. For example, inspired by congestion control algorithms, TwitInfo \cite{marcus2011twitinfo} used a weighted moving average and variance model to detect peaks of terms in Twitter data to track an event. 
2) Probabilistic topic modeling. For example, Latent Event and Category Model (LECM) \cite{zhou2015unsupervised,cai2015popular} modeled each tweet as a joint distribution over a range of features (e.g., text, image, named entities, time, location, etc.). 
3) Unsupervised clustering approaches. New tweets are determined to merge into an existing cluster or form a new cluster based on a threshold of similarity \cite{becker2011beyond}, and events are summarized from clusters using metrics such as popularity, freshness and confidence scores.




Our event recognition approach is closely related to supervised classification approaches for Twitter event detection \cite{zhang2019social}.   
Different classification methods, Naive Bayes \cite{sankaranarayanan2009twitterstand}, Support Vector Machines \cite{sakaki2010earthquake}, Decision Trees \cite{popescu2011extracting}, and Neural Networks \cite{caragea2016identifying,nguyen2017robust} have been used to train event recognizers using human annotated Twitter messages. However, annotating a large number of Twitter messages for a new disaster is 
time-consuming, 
and systems trained using old labeled data  
may be biased to only detect information specific to one historical disaster (e.g., local road names, local authorities, etc.). 
In contrast, 
the weakly supervised classification 
approach we propose does not 
require slow brewed training instances annotated one by one, and can quickly label data and train event recognizers from scratch for a newly happened disaster.

\begin{figure*}[t]
 \begin{center}
  \includegraphics[width = 5.3in]{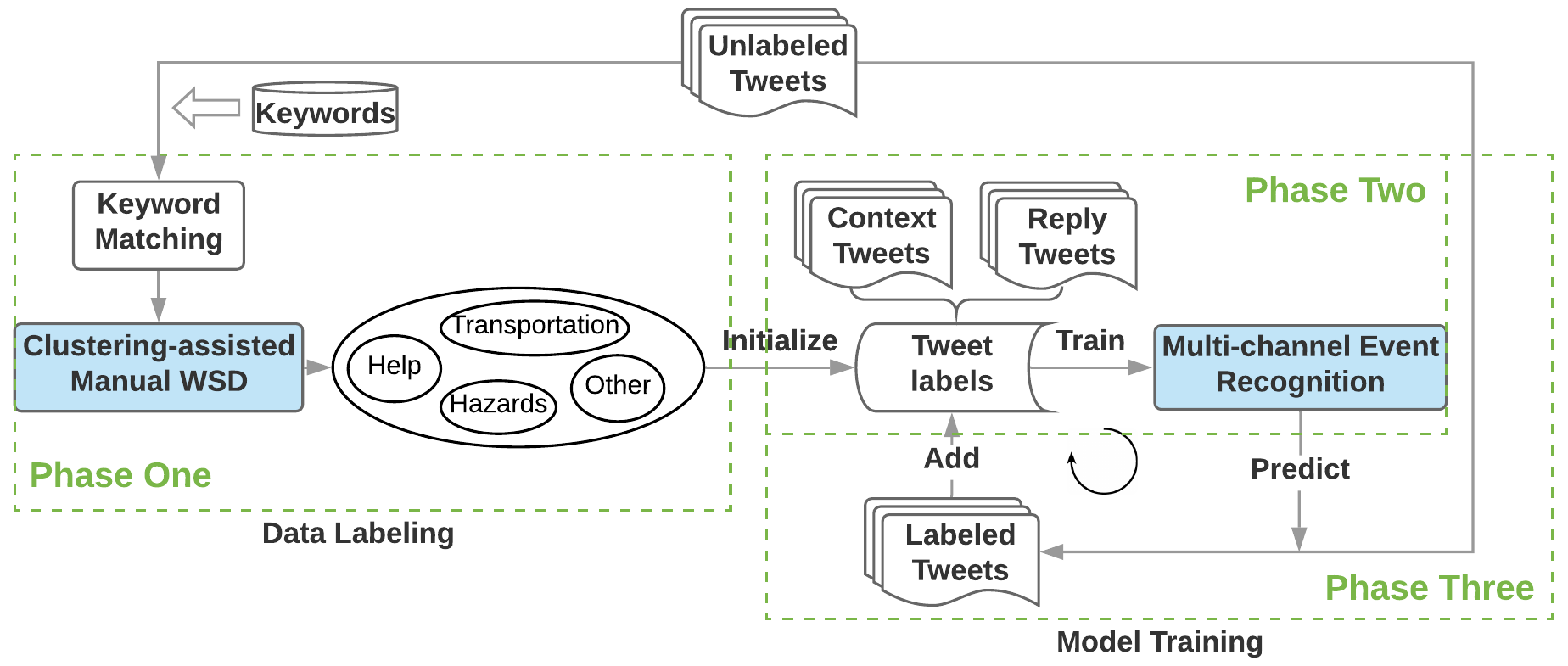}
  \end{center}
 \caption{Overview of the Weakly-supervised Learning System}
\label{flow_chart}
\end{figure*}

\section{Event Categories and Event Keywords}\label{EventCategories}
Disaster management generally consists of four phases - mitigation, preparedness, response, and recovery. 
We focus on identifying events during the {\it response} phase of disasters, which is arguably the most crucial and time-critical part of emergency management. 
Based on an existing event ontology for hurricanes \cite{huang2015geographic}, we 
identified nine types of events, including three types of human activity events and six types of built environment related events, as briefly described below.

\noindent{\bf Human activities}. 
1) Preventative measure (PRE). 
People look for shelters or process evacuation; 
Any flood-proof processes (e.g., building waterproof facilities, etc.).
2) Help and rescue (RES). People provide, receive, or seek face to face help in disastrous environments, including indirect help such as donating money, supply, and providing services.
3) Casualty (CAS). Disaster-caused death, injury, hurt, etc.

\noindent{\bf Built environment}.
4) Housing (HOU). Reporting emergencies of a house, apartment, home, etc.
5) Utilities and Supplies (UTI). Problems with heating, gas, water, power, communication facility, food, grocery stores, etc.
6) Transportation (TRA). The impact on the traffic, bus services, or the closure of a road, airport, highway, etc.
7) Flood control infrastructures (FCI). The impact on or damage to the reservoir, bayou, canal, dam, etc.
8) Business, Work, School (BWS). The changes of schedule, e.g., business closed/open, school closed/open, etc.
9) Built-environment hazards (HAZ). The damage or risks that may cause injury or death related to the built environment, such as fire, explosion, contamination, electric shock, debris, etc.

Meanwhile, 
the event ontology \cite{huang2015geographic} contains event keywords, 
and we selected at most five keywords 
for each event category that are not specific to any particular hurricane or location, 
e.g., keywords ``evacuate'' and ``shelter'' for the category of Preventive measure (PRE), and ``help'' and ``rescue'' for Help and rescue (RES), etc. 

\section{Our Approach}
Figure \ref{flow_chart} gives an overview of our  weakly-supervised learning approach with three phases.
In phase one, we quickly create high-quality labeled data. 
Specifically, we conduct clustering-assisted manual word sense disambiguation on event keyword identified noisy tweets, to significantly clean and improve the quality
of automatically labeled tweets. 
In phase two, we train a multi-channel BiLSTM classifier using tweets  
together with their context tweets and reply tweets. In phase three,  we iteratively retrain the multi-channel classifier to further improve its event recognition performance. 

\subsection{Phase One: Rapid Data Labeling via Clustering Assisted Manual Word Sense Disambiguation}

For each event category, we first retrieve tweets containing a predefined event keyword and then apply a clustering algorithm to form tweet clusters. 
To facilitate manual word sense disambiguation, we rank tweet clusters based on their sizes (number of tweets) and then ask a domain expert to judge whether a cluster (from largest to smallest) shows the pertinent meaning of an event keyword by inspecting five example tweets randomly sampled from the cluster. 
The annotator stops scrutiny once 
20 pertinent clusters are identified for each event category\footnote{If a cluster adopts a word sense that is irrelevant to any event category, we assign the catch-all label ``Other'' to it and later use tweets in such clusters as negative training instances.}. 
 After cleaning, around a third to half
 of keyword identified tweets were removed. Specifically, 6.6K out of 15.2K keyword identified tweets and 5.8K out of 17.5K keyword identified tweets were removed in the {\it Harvey} and {\it Florence} datasets respectively.
 
Next, we describe the clustering algorithm used here, the Speaker-Listener Label Propagation Algorithm (SLPA) \cite{xie2011slpa}. 

\subsubsection{The Clustering Algorithm}
\label{clustering_alg}
The SLPA algorithm 
is initially introduced to discover overlapping communities in social user networks, where one user may belong to multiple communities. 
The basic idea of SLPA is to simulate people's behavior of spreading the most frequently discussed topics among neighbors. 
We choose SLPA for two reasons. First, SLPA is a self-adaptation model that can automatically converge to the optimal number of communities, so no pre-defined number of communities is needed. 
Second, a tweet during natural disasters may mention more than one event, which corresponds to one user  belonging to multiple communities. SLPA has been shown one of the best algorithms for detecting overlapping communities \cite{xie2013overlapping}. 

{\bf Clustering with Graph Propagation:}  SLPA is essentially an iterative algorithm. It first initializes each node as a cluster by itself. In listener-speaker propagation iterations, each node will be chosen in turn to be either a listener or a speaker. 
Each time, a listener node accepts the label that is the most popular among its neighbors 
and accumulates such knowledge in the memory. And a speaker advocates one label based on the probability distribution updated in its memory. 
Finally, based on the memories, connected nodes sharing a label with the probability over a threshold are grouped together and form a community.

We modified the original SLPA to make it suitable for clustering Twitter messages. Formally, given a set of tweets, we construct an undirected graph $G(V, E)$, where $V$ represents all tweets and $E$ represents weighted edges between nodes. 
The weight of an edge $e$ between two tweets $u$ and $v$ is calculated based on content similarity of the two tweets. In label propagation, we consider weighted voting to determine the cluster of a tweet. 

{\bf The Similarity Measure:}
Determining similarities between nodes is important for clustering algorithms. 
However, Twitter messages are informal 
and often contain meaningless words, 
therefore, we aim to first select important words before calculating content similarities between tweets. 
Recently, \cite{conneau2017supervised} 
proposed an approach for learning universal sentence representations using the Stanford Natural Language Inference (SNLI) dataset \cite{snli:emnlp2015} and demonstrated its effectiveness in reasoning about semantic relations 
between sentences. We notice that Twitter messages and SNLI data have two common characteristics: short sentences in a casual language. Hence, we apply their learned sentence representation constructor to tweets for identifying important words. 

Specifically, 
for a given tweet with $T$ words $\{w_t\}_{t=1,2,...,T}$, we applied the pre-trained Bi-directional LSTMs \cite{conneau2017supervised} to compute $T$ hidden vectors,  $\{h_t\}_{t=1,2,...,T}$, one per word. 
Next, for each dimension, we determine the maximum value over all the hidden vectors $\{h_t\}_{t=1,2,...,T}$. The importance score for a word $w_t$ is calculated as the number of dimensions where its hidden vector $h_t$ has the maximum value divided by the total number of dimensions. 
Then, we select words having importance scores $\geq$ the average importance score (1.0 / the number of words) as important words. 
For example, in the following tweet, the bolded words are selected:
{\it It has \textbf{started} a \textbf{fundraiser} for \textbf{hurricane} Harvey \textbf{recovery} efforts in \textbf{Houston}, you can \textbf{donate} here.}

We calculate the similarity score between two tweets by   considering only selected words shared by two tweets.  Empirically, we found this similarity measure 
performs better than the straightforward cosine similarity measure considering all words. 
Specifically, the similarity score between two tweets $u$ and $v$ is the number of common words  / (length of $u$ $\times$ length of $v$). To construct the tweet graph, we create an edge between two tweets when they share two or more selected words and the edge weight is their similarity score.


\subsection{Phase Two: Multi-channel Tweet Classification}
The most unique characteristic of social media is the network structure which not only connects users (e.g., friend network or follower network), but also makes Twitter messages connected. Therefore, we exploit other related tweets for enhancing the representation of a target tweet.  
In particular, we found the immediately preceding context tweets and reply tweets useful. 

First, the past tweets written by the same user provide additional evidence for an event recognition system to infer the event topic of the current tweet. 
Interestingly, we observe that 
the event topic 
engaging a user's attention is usually consistent within a small time window, as shown in the upper example of Figure \ref{context_reply} where the two relevant context tweets are within 2 minutes. 
We further observe that the topic relatedness between the target tweet and context tweets decreases 
quickly over time. 
In our experiments, we only consider a relatively small number of context tweets, specifically five of the preceding tweets.  
In addition, we 
assign a weight to a context tweet as $w_i = 0.8^{m_i}$, where $m_i$ is the time distance (in minutes) between the $i^{th}$ context tweet and the target tweet. 

Second, 
reply tweets usually provide information that is hidden in the original tweet, as shown in the lower example of Figure \ref{context_reply}. But compared to regular Twitter posts, replies are much noisier. To select the most informative reply tweets for a given target tweet, we rank replies according to the number of common words they share with the target tweet and pick a small number of them from the top, specifically at most five replies.

\begin{figure}[t]
 \begin{center}
 \includegraphics[width = 2.4in]{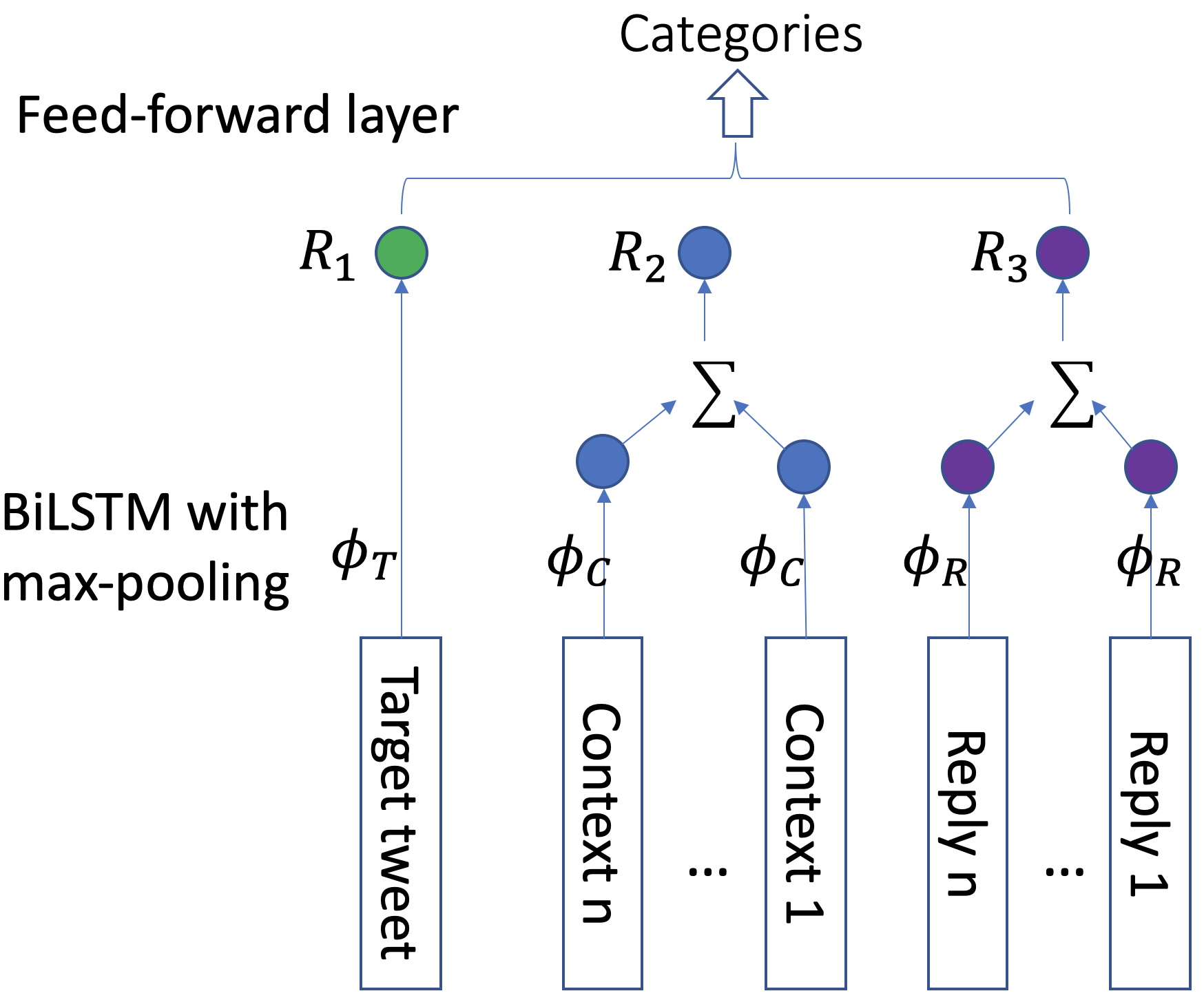}
  \end{center}
  \setlength{\belowcaptionskip}{-10pt}
 \caption{BiLSTM Classifier using Context and Reply Enriched Representation}
\label{BiLSTM}
\end{figure}

Figure \ref{BiLSTM} shows the overall structure of the classifier. 
\begin{equation}
\label{concatenation}
\begin{aligned}
R_{1} &= \Phi_T(tw^{target})\\
R_{2} &= \frac{1}{\sum w_i}\sum_{i=1}^{N}w_i\cdot \Phi_C(tw_{i}^{context})\\
R_{3} &= \frac{1}{M} \sum_{i=1}^{M}\Phi_R(tw_{i}^{reply})\\
R_{all} &= [R_1, R_2, R_3]
\end{aligned}
\end{equation}
Specifically, 
we apply three separate BiLSTM encoders \cite{graves2005framewise} with max-pooling \cite{collobert2008unified} 
 to obtain sentence embeddings for the target tweet, context tweets and reply tweets (i.e., $\Phi_T$, $\Phi_C$, $\Phi_R$). 
Then, the 
final enriched representation of the target tweet ($R_{all}$) is the concatenation of the target tweet embedding ($R_{1}$),  
weighted average of context tweet embeddings ($R_{2}$),  
and unweighted average of reply tweet embeddings ($R_{3}$). 

On top of $R_{all}$, we apply a feedforward neural net to directly map $R_{all}$ to 10 classes (9 event categories + Other). 
We 
optimizes a multi-label one-versus-all loss based on max-entropy, considering that one tweet may belong to multiple event categories.  
To deal with imbalanced distributions of event categories, we re-scale the prediction loss of each class (proportional to $\frac{1}{ClassSize}$) so that smaller classes are weighted more heavily in the final loss function. 
For all BiLSTM encoders, we use one hidden-layer of 300 units, pre-trained GloVe \cite{pennington2014glove} word embeddings of 300 dimensions, Adam optimizer \cite{kingma2014adam} with a learning rate of 0.0001.

In training, to compete with positive training instances (tweets labeled with any event category), we randomly sample unlabeled tweets equal to the sum of labeled tweets in size   
and use them as negative training instances (the category Other), 
to reflect the fact that there are generally more tweets reporting no event. 


\begin{table*}[t]
\small
\begin{center}

\begin{tabular}{l|cccccccccc|c}

Category & PRE & RES & CAS & HOU & UTI & TRA & FCI & BWS & HAZ & Other & Sum \\ \hline
\multicolumn{12}{c}{Harvey (Aug.28 1:00-2:00 pm)}\\\hline
Amount & 374 & 1092 & 43 & 142 & 270 & 225 & 73 & 501 & 30 & 9165 & 11782\\
Percentage & 3.2\% & 9.3\% & 0.4\% & 1.2\% & 2.3\% & 1.9\% & 0.6\% & 4.3\% & 0.3\% & 77.8\% & 100\% \\\hline
\multicolumn{12}{c}{Florence (Sept.17 1:00-1:30 pm)}\\\hline
Amount & 69 & 490 & 120 & 28 & 146 & 85 & 8 & 80 & 23 & 3031 & 4059\\
Percentage & 1.7\% & 12.1\% & 3\% & 0.7\% & 3.6\% & 2.1\% & 0.2\% & 2\% & 0.6\% & 74.7\% & 100\% \\\hline
\end{tabular}
\end{center}
\caption{Annotation: Number of Tweets in Each Event Category}
\label{tweets_distribution}
\end{table*}

\subsection{Phase Three: Improve Coverage with Bootstrapping Learning}

After the first two phases,  
we have 
labeled tweets by conducting time-efficient clustering-assisted WSD on event keyword identified tweets and have used these quickly labeled tweets to train the multi-channel event recognizer. However, all the labeled tweets yielded in phase 1 contain a predefined event keyword, while 
 many event categories may have cases that 
 do not contain a keyword. 
 Therefore, 
we further exploit bootstrapping learning and iteratively improve the coverage of the multi-channel classifier. 

Specifically, we apply the initial multi-channel classifier on  unlabeled tweets and label new tweets for each event category.
Newly labeled tweets together with their context tweets and replies are used to retrain the model. 
To enforce the classifier to look at new content words other than  event keywords,  
we randomly cover 20\% of keywords 
occurrences in every training epoch, inspired by \cite{srivastava2014dropout}.  
In order to combat semantic drifts \cite{mcintosh2009reducing} in bootstrapping learning, we initially apply 
a high confidence score for selecting newly labeled tweets used to retrain the classifier and lower the confidence score  gradually. 
Specifically, the confidence score was initially set at 0.9 and lowered by 0.1 each time when the number of selected tweets is less than 100. The bootstrapping process stops when the confidence score decreases to 0.5\footnote{In our experiment, the bootstrapping process stopped after 9 iterations, and each iteration took around 10 minutes using a NVIDIA's GeForce GTX 1080 GPU.}.

\begin{table*}[t]
\small
\begin{center}
\begin{tabular}{ll|ccccccccc|c}
 Row & Method & PRE & RES & CAS & HOU & UTI & TRA & FCI & BWS & HAZ & Macro Average \\\hline

1 & Keyword Matching &  73.9 &  56.6 &  26.2 &  36.4 &  54.3 &  38.0 &  54.4 &  55.5 &  43.1 &  51.1/52.5/51.8 \\

2 & LDA &  39.4 &  41.3 &  4.9 &  8.5 &  19.8 &  28.8 &  40.4 &  17.7 &  25.5 &  19.6/42.6/26.8 \\ 

3 & Guided LDA &  43.4 &  45.8 &  10.1 &  8.6 &  21.1 &  40.7 &  53.4 &  20.4 &  24.5 &  25.1/45.2/32.3 \\ 

4 & SLPA &  61.4 &  61.3 &  18.9 &  23.1 &  36.4 &  36.2 &  56.5 &  44.4 &  23.3 &  39.6/48.1/43.4 \\


 \hline

\multicolumn{12}{c}{\bf Seed with Keyword Identified Tweets with no Cleaning }    \\ \hline

5 & Basic Classifier &  82.6 &  63.8 &  18.8 &  36.9 &  60.2 &  36.8 &  61.7 &  61.0 &  45.5 &  50.3/60.5/54.9 \\ 

6 & \hspace{0.2cm}  + bootstrapping &  82.6 &  64.1 &  20.1 &  37.3 &  60.6 &  36.5 &  62.8 &  60.6 &  45.7 &  50.2/61.2/55.3 \\ 

7 & Multi-channel Classifier &  {\bf 84.3} &  68.6 &  22.1 &  37.1 &  60.5 &  40.5 &  62.3 &  62.1 &  45.7 &  50.5/64.1/56.5 \\ 

8 & \hspace{0.2cm}  + bootstrapping & 84.1 &  69.1 &  22.6 & 36.4 & 59.6 &  42.1 &  63.1 & 60.4 & 46.4 &  49.4/{\bf 65.8}/56.4 \\ \hline


\multicolumn{12}{c}{\bf Seed with Keyword Identified Tweets Cleaned by Clustering-assisted WSD}    \\ \hline

9 & Basic Classifier &  82.6 &  68.4 &  34.4 &  45.1 &  65.8 &  56.4 &  63.3 &  68.4 &  49.0 &  68.3/57.2/62.3 \\ 

10 & \hspace{0.2cm}  + bootstrapping &  83.5 &  68.8 &  36.9 &  45.0 &  65.8 &  58.4 &  66.7 &  68.5 &  51.9 &  67.1/59.9/63.3 \\ 



11 & Multi-channel Classifier &  82.8 &  68.0 &  36.7 &  {\bf 47.6} &  65.1 &  57.2 &  63.3 &  67.8 &  56.5 &  72.5/57.0/63.8 \\ 

12 & \hspace{0.2cm}  + bootstrapping &  83.9 &  67.8 &  36.7 &  45.7 &  {\bf 66.1} &  61.3 &  {\bf 74.8} &  69.1 &  {\bf 57.7} &  70.1/61.6/{\bf 65.5} \\ \hline
\hline
13 & Supervised Classifier &  80.8 &  {\bf 72.2} &  {\bf 48.0} &  45.3 &  56.3 &   {\bf 67.9} &  65.9 &  {\bf 71.2} &  45.3 &  {\bf 73.2}/53.6/61.9 \\ \hline

\end{tabular}
\end{center}
\setlength{\belowcaptionskip}{-10pt}
\caption{Experimental Results on Hurricane Harvey: F1-score for each event category and macro-average Precision/Recall/F1-score (\%) over all categories.} 
\label{harvey_performance}
\end{table*}

\section{Experiments and Results}
\subsection{Data Sets}

We apply the approach to datasets for two hurricanes, {\it Harvey}  (the primary dataset) and {\it Florence} (the second dataset). 
Hurricane {\it Harvey} struck the Houston metropolitan area and Southeast Texas 
in 2017, and ranks as the second costliest hurricane (\$125 billion in damage) on record for the United States \cite{HarveyImpact}. Hurricane Florence also caused severe damage (more than \$24 billion) in the North and South Carolina 
in 2018.
To retrieve tweets in affected areas, we consider two constraints in twitter crawling using GNIP API \cite{GNIP}: 1) a tweet has the geo-location within affected areas (Houston or major cities in Carolinas) or 2) the author of a tweet has his/her profile located in affected areas. 
Since we aim to 
recognize original tweet messages reporting events for disaster management purposes, we only consider original tweets as target tweets for classifications across all the experiments and we ignore retweets and reply tweets. 

To create the official evaluation data (details in the next section), 
we exhaustively annotated all the tweets posted from 1:00 to 2:00 pm, August 28, 2017 for {\it Harvey} and from 1:00 to 1:30 pm, September 17, 2018 for {\it Florence}, both among the most impacted time periods for the two hurricanes. For training both our systems and the baseline systems, we used around 65k and 69.8k  unlabeled tweets for {\it Harvey} and {\it Florence} respectively that were posted 12 hours (half a day) preceding the test time period and are therefore strictly separated from the tweets used for evaluation. 

\subsection{Human Annotations for Evaluation}\label{anno}

In order to obtain high-quality evaluation data, 
we trained two annotators and refined annotation guidelines for several rounds. 
A tweet is annotated with an event category if it directly 
discusses events of the defined category, including sharing information and expressing opinions. 
A tweet may receive multiple labels if it discusses more than one event and the events are of different types.
If one tweet does not discuss any event of an interested type, we label it as {\it Other}. 

 We first asked the two annotators to annotate a common set of 600 tweets from the {\it Harvey} set and 
 they achieved a substantial kappa score of 0.67 \cite{cohen1968weighted}. We then split the remaining annotations evenly between the two annotators. 
The distributions of annotated tweets 
are shown in Table \ref{tweets_distribution}.\footnote{A small number of tweets were annotated with more than one event category, 290 (11\%) and 57 (6\%) tweets for {\it Harvey} and {\it Florence} datasets respectively.} 
Consistent across the two considered hurricane disasters, tweets describing interested events cover only around one quarter of all posted tweets and their distributions over the event categories are highly imbalanced. 



\subsection{Unsupervised Baseline Systems}\label{baselines}

\noindent{\bf Keyword matching:}
labels a tweet with an event category if the tweet contains any keyword in the event category. 
A tweet may be assigned to multiple event categories if the tweet contains keywords from more than one event category.

\noindent{\bf Topic modeling Approaches:}
Probabilistic topic modeling approaches have been commonly used to identify latent topics from a collection of documents. 
We assign each topic to an event category if the top ten words of a topic ranked by word probabilities contain any keyword of the  category. 
A topic may be assigned to multiple event categories if its top ten words contain keywords from more than one category.
Given a new tweet, we infer its topics and assign the event labels of the most significant topic.  
We implement 
two topic modeling approaches. 
{\bf LDA (Latent Dirichlet Allocation)} \cite{blei2003latent} assumes a document can be represented as a mixture over latent topics, where each topic is a probabilistic distribution over words. 
{\bf Guided LDA} \cite{jagarlamudi2012incorporating} 
is a stronger version of LDA, that incorporates our predefined event keywords to guide the topic discovery process. 
For fair comparisons, we also apply important words selection used in our system 
for LDA and GuidedLDA\footnote{We also tried LDA and Guided LDA without important words selection which yields a much worse performance.}. Note that 
both approaches require pre-defining the number of topics, which is hard to estimate, we 
set this hyper-parameter as 100 
in our experiments.

\noindent{\bf SLPA:}
We also apply the adapted SLPA clustering algorithm to form clusters and assign each cluster to an event category if the top ten words in a cluster ranked by word frequencies contain any keyword of the category. 
Given a new tweet, we identify its neighbor tweets using the same similarity measure we used for clustering in phase one 
and label the tweet with the majority event label over its neighbors. 



\subsection{Results on Hurricane {\it Harvey}}
Table \ref{harvey_performance} shows the experimental results. 
The first section shows performance of 
baseline systems.  
Among the four baselines, the simple keyword matching approach (row 1) performs the best, 
and the clustering algorithm SLPA (row 4) outperforms both LDA-based approaches (row 2 \& 3). 
The event recognition performance of these mostly unsupervised systems is consistently low, presumably due to their incapability to resolve severe lexical ambiguities in tweets.




The second section of Table \ref{harvey_performance} shows results of four classifiers that directly use keyword identified noisy tweets with no cleaning for training. 
Row 5 shows the results of the basic classifier considering the target tweet only. Row 7 shows the results of the multi-channel classifier that further considers contexts and replies, which yields a small recall gain compared to row 5. 
Row 6 \& 8 show the results of the two classifiers after applying bootstrapping learning, which further improves the recall a bit. However, the precision of all the four classifiers is 
around 50\% similar to the keyword matching baseline and  
consistently unsatisfactory.

The third section of Table \ref{harvey_performance} shows results of the same set of classifiers but using clustering-assisted WSD cleaned tweets for training. 
Compared to its counterpart 
 trained using noisy tweets (row 5), 
the precision of the basic classifier (row 9) improves significantly by 18\%. With a small drop on recall, the overall F-score improves by 7.4\%. 
The multi-channel classifier (row 11) further improves the precision with an almost identical recall. Bootstrapping learning 
 improves the recall of both classifiers. The full system (row 12) outperforms its counterpart 
 trained using noisy tweets 
 (row 8) by over 20\% in precision and 9\% in F-score. 
 Meanwhile, using a little supervision, the rapidly trained weakly supervised system greatly outperforms the unsupervised baseline systems, 
 yielding 20\% (or more) and 15\% (or more) of increases in precision and F-score respectively. 

\begin{figure}[t]
 \begin{center}
 \includegraphics[width = 2.5in]{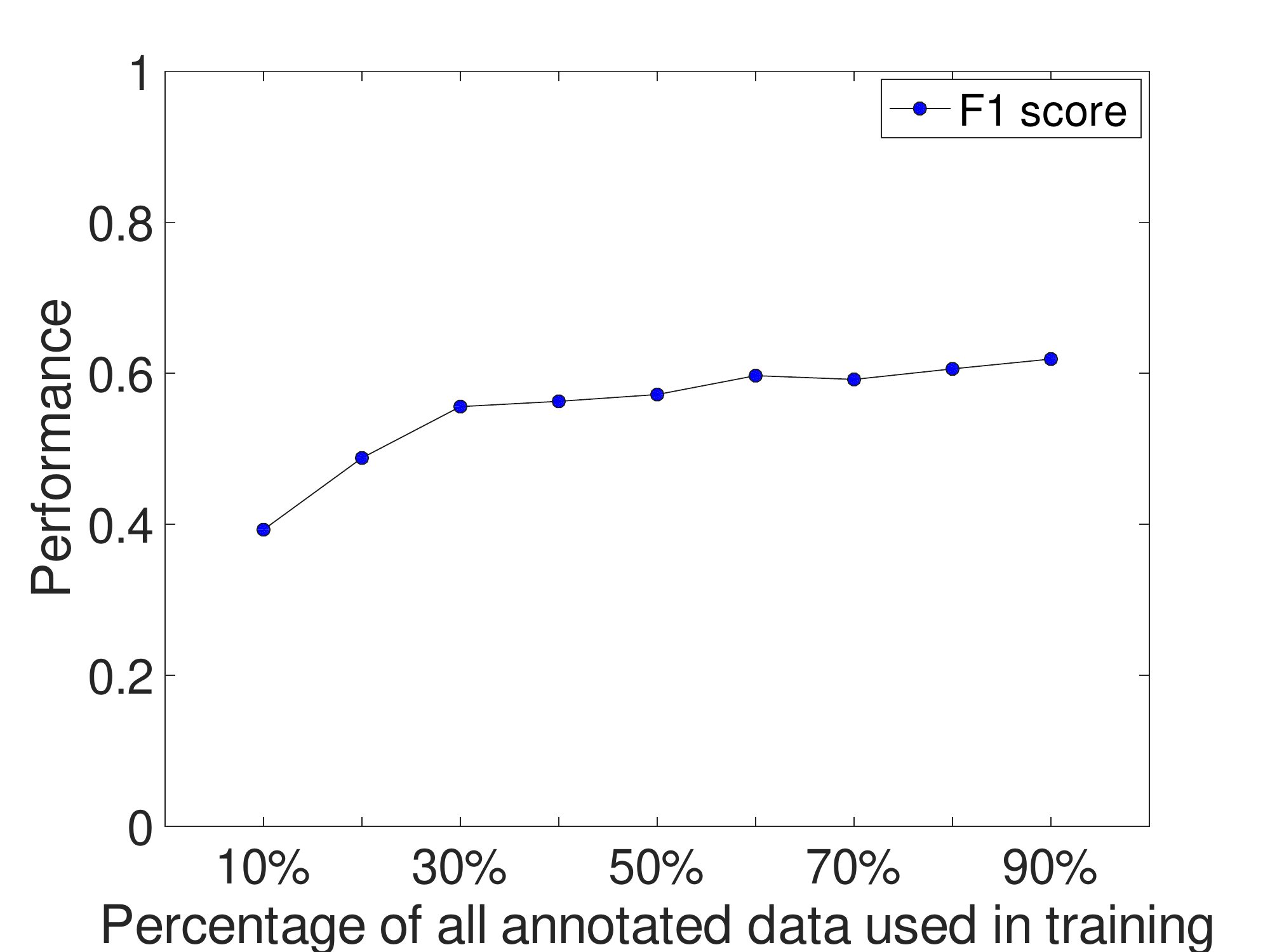}
  \end{center}
 \caption{Learning curve of 10-fold cross validation}
\label{learning_curve}
\end{figure}

\noindent{\bf Comparisons with Supervised Learning:} 
We train and evaluate a supervised classifier (multi-channel) using annotated tweets under the 10-fold cross validation setting.   
 The results of the supervised classifier are shown in the last row of Table  \ref{harvey_performance}.  
Compared to the supervised classifier, 
the weakly supervised approach yields a recall gain of 8\% with a slightly lower precision, and improves the overall F-score by 3.6\%. 
Note that around 50 person-hours were needed to 
annotate over 11K tweets following the normal tweet-by-tweet annotation process, 
while our data labeling method 
only required 1-2 person-hours for clustering-assisted WSD. 
Considering that 
a large number of 
tweets are 
time-consuming to annotate, we conducted another group of experiments that gradually add 
annotations in training to see how the size of training data affects the performance.  Specifically, under 10-fold cross validation, we randomly sample a certain percentage of tweets from nine training folds as training data, ranging from
0.1 to 0.9 in increments of 0.1. 
The learning curve 
(Figure \ref{learning_curve}) 
is steep in the beginning and then levels out as the remaining 70\% of 
annotated tweets (around 7K tweets) were continuously appended, which shows that the normal annotation method may create many redundant annotations. 

\begin{table}[t]
\small
\begin{center}
\begin{tabular}{l|c}
 & Macro Average \\\hline

Keywords &  43.7/46.9/45.3 \\ \hline

\multicolumn{2}{c}{\bf with no Cleaning}    \\ \hline

Basic Classifier &  40.8/47.9/44.1  \\ 

\hspace{0.2cm}  + bootstrapping &  39.8/52.8/45.4  \\ 

Multi-channel Classifier &  43.1/48.7/45.8 \\ 

\hspace{0.2cm}  + bootstrapping &   41.2/52.6/46.2 \\ \hline

\multicolumn{2}{c}{\bf with Clustering-assisted WSD}    \\ \hline

Basic Classifier &  67.8/49.6/57.3  \\ 

\hspace{0.2cm}  + bootstrapping &  63.4/54.9/58.8  \\ 

Multi-channel Classifier &  {\bf 70.3}/50.2/58.5 \\ 

\hspace{0.2cm}  + bootstrapping &   65.1/{\bf 55.1}/{\bf 59.7} \\ \hline
\hline
Supervised Classifier & 57.8/40.9/47.9\\
\hline

\end{tabular}
\end{center}
\caption{Experimental Results on Hurricane Florence (Precision/Recall/F1-score \%)}
\label{florence_performance}
\end{table}
\subsection{Results on Hurricane {\it Florence}}
Table \ref{florence_performance} shows the results. Similar to 
Hurricane {\it Harvey},   
clustering-assisted WSD clearly improves the precision of the trained classifier for Hurricane {\it Florence} as well.    Enriching tweet representations and conducting bootstrapping learning further improve the performance of the full system, which clearly outperforms the supervised classifier.   

\subsection{Analysis}
\begin{figure}[t]
 \begin{center}
 \includegraphics[width = 2.3in]{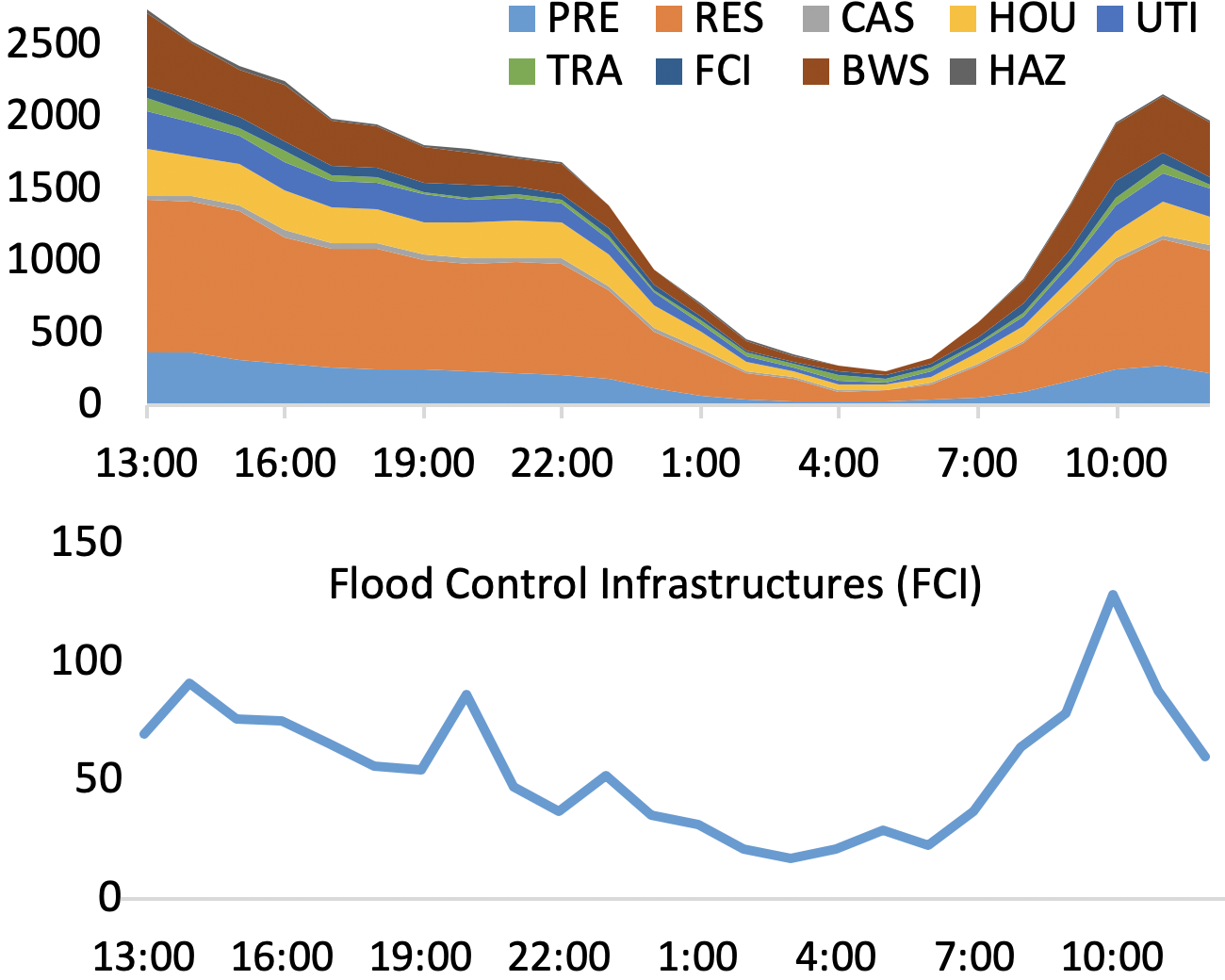}
  \end{center}
 \caption{Curves for all the categories (Upper) and for Flood Control Infrastructures only (Lower).}
\label{event_trend}
\end{figure}

For Hurricane {\it Harvey}, we applied the 
full system to label tweets posted right after the test hour. 
Figure \ref{event_trend} plots the number of tweets detected for each hour. 
Overall, the clear low point corresponds with the day-night shift. Taking a closer look at the curve for the flood control infrastructure category, we can see an obvious burst at 8 pm  Aug.28, 2017, 
triggered by an official update 
on water release of two major reservoirs, as well as a burst at 10 am Aug.29, 
triggered by the collapse of a bridge over Greens Bayou, with example tweets shown in Figure \ref{fig:narrative_example}. 

\begin{figure}[t]
\fbox{\begin{minipage}{23em}
\small
HAPPENING NOW: @hcfcd live update on Addicks Reservoir and certain levees. Watch now on TV or here.\\
One of the dams they want to discharge is near me.
\end{minipage}}
\fbox{\begin{minipage}{23em}
\small
BREAKING: The levee at Columbia Lakes has been breached! GET OUT NOW! PLEASE BE SAFE!\\
A bridge has collapsed at Greens Bayou. Be careful!
\end{minipage}}
\caption{Example tweets sampled from two bursts}
\label{fig:narrative_example}
\end{figure}

\section{Conclusion}

We have presented a weakly supervised event recognition system that can effectively recognize fine-grained event categories for  individual tweet messages. 
We highlight the novel clustering-assisted manual word sense disambiguation data labeling method that is time-efficient and significantly improves the quality of event keyword identified texts. 
The evaluation on two hurricanes show the effectiveness and robustness of the overall approach.  
The weakly supervised system can be easily adapted to other  disaster types (e.g., earthquake, tsunami, etc.) with a relevant event ontology to support real-time disaster management. 

\section{Acknowledgments}
We gratefully acknowledge support from National Science Foundation via the awards IIS-1759537 and IIS-1755943.

\newpage

\bibliography{my_aaai}

\begin{thebibliography}{}

\bibitem[\protect\citeauthoryear{Becker, Naaman, and
  Gravano}{2011}]{becker2011beyond}
Becker, H.; Naaman, M.; and Gravano, L.
\newblock 2011.
\newblock Beyond trending topics: Real-world event identification on twitter.
\newblock {\em Icwsm} 11(2011):438--441.

\bibitem[\protect\citeauthoryear{Blei, Ng, and Jordan}{2003}]{blei2003latent}
Blei, D.~M.; Ng, A.~Y.; and Jordan, M.~I.
\newblock 2003.
\newblock Latent dirichlet allocation.
\newblock {\em Journal of machine Learning research} 3(Jan):993--1022.

\bibitem[\protect\citeauthoryear{Bowman \bgroup et al\mbox.\egroup
  }{2015}]{snli:emnlp2015}
Bowman, S.~R.; Angeli, G.; Potts, C.; and Manning, C.~D.
\newblock 2015.
\newblock A large annotated corpus for learning natural language inference.
\newblock In {\em Proceedings of the 2015 Conference on Empirical Methods in
  Natural Language Processing (EMNLP)}.
\newblock Association for Computational Linguistics.

\bibitem[\protect\citeauthoryear{Cai \bgroup et al\mbox.\egroup
  }{2015}]{cai2015popular}
Cai, H.; Yang, Y.; Li, X.; and Huang, Z.
\newblock 2015.
\newblock What are popular: exploring twitter features for event detection,
  tracking and visualization.
\newblock In {\em Proceedings of the 23rd ACM international conference on
  Multimedia},  89--98.
\newblock ACM.

\bibitem[\protect\citeauthoryear{Caragea, Silvescu, and
  Tapia}{2016}]{caragea2016identifying}
Caragea, C.; Silvescu, A.; and Tapia, A.~H.
\newblock 2016.
\newblock Identifying informative messages in disaster events using
  convolutional neural networks.
\newblock In {\em International Conference on Information Systems for Crisis
  Response and Management},  137--147.

\bibitem[\protect\citeauthoryear{Cohen}{1968}]{cohen1968weighted}
Cohen, J.
\newblock 1968.
\newblock Weighted kappa: Nominal scale agreement provision for scaled
  disagreement or partial credit.
\newblock {\em Psychological bulletin} 70(4):213.

\bibitem[\protect\citeauthoryear{Collobert and
  Weston}{2008}]{collobert2008unified}
Collobert, R., and Weston, J.
\newblock 2008.
\newblock A unified architecture for natural language processing: Deep neural
  networks with multitask learning.
\newblock In {\em Proceedings of the 25th international conference on Machine
  learning},  160--167.
\newblock ACM.

\bibitem[\protect\citeauthoryear{Conneau \bgroup et al\mbox.\egroup
  }{2017}]{conneau2017supervised}
Conneau, A.; Kiela, D.; Schwenk, H.; Barrault, L.; and Bordes, A.
\newblock 2017.
\newblock Supervised learning of universal sentence representations from
  natural language inference data.
\newblock In {\em Proceedings of the 2017 Conference on Empirical Methods in
  Natural Language Processing},  670--680.

\bibitem[\protect\citeauthoryear{Graves and
  Schmidhuber}{2005}]{graves2005framewise}
Graves, A., and Schmidhuber, J.
\newblock 2005.
\newblock Framewise phoneme classification with bidirectional lstm and other
  neural network architectures.
\newblock {\em Neural Networks} 18(5-6):602--610.

\bibitem[\protect\citeauthoryear{Huang and Xiao}{2015}]{huang2015geographic}
Huang, Q., and Xiao, Y.
\newblock 2015.
\newblock Geographic situational awareness: mining tweets for disaster
  preparedness, emergency response, impact, and recovery.
\newblock {\em ISPRS International Journal of Geo-Information} 4(3):1549--1568.

\bibitem[\protect\citeauthoryear{Jagarlamudi, Daum{\'e}~III, and
  Udupa}{2012}]{jagarlamudi2012incorporating}
Jagarlamudi, J.; Daum{\'e}~III, H.; and Udupa, R.
\newblock 2012.
\newblock Incorporating lexical priors into topic models.
\newblock In {\em Proceedings of the 13th Conference of the European Chapter of
  the Association for Computational Linguistics},  204--213.
\newblock Association for Computational Linguistics.

\bibitem[\protect\citeauthoryear{Kingma and Ba}{2014}]{kingma2014adam}
Kingma, D.~P., and Ba, J.
\newblock 2014.
\newblock Adam: A method for stochastic optimization.
\newblock {\em arXiv preprint arXiv:1412.6980}.

\bibitem[\protect\citeauthoryear{Marcus \bgroup et al\mbox.\egroup
  }{2011}]{marcus2011twitinfo}
Marcus, A.; Bernstein, M.~S.; Badar, O.; Karger, D.~R.; Madden, S.; and Miller,
  R.~C.
\newblock 2011.
\newblock Twitinfo: aggregating and visualizing microblogs for event
  exploration.
\newblock In {\em Proceedings of the SIGCHI conference on Human factors in
  computing systems},  227--236.
\newblock ACM.

\bibitem[\protect\citeauthoryear{McIntosh and
  Curran}{2009}]{mcintosh2009reducing}
McIntosh, T., and Curran, J.~R.
\newblock 2009.
\newblock Reducing semantic drift with bagging and distributional similarity.
\newblock In {\em Proceedings of the Joint Conference of the 47th Annual
  Meeting of the ACL and the 4th International Joint Conference on Natural
  Language Processing of the AFNLP},  396--404.

\bibitem[\protect\citeauthoryear{{National Hurricane
  Center}}{2017}]{HarveyImpact}
{National Hurricane Center}.
\newblock 2017.
\newblock Costliest u.s. tropical cyclones tables updated.
\newblock Technical report.

\bibitem[\protect\citeauthoryear{Navigli and
  Lapata}{2010}]{navigli2010experimental}
Navigli, R., and Lapata, M.
\newblock 2010.
\newblock An experimental study of graph connectivity for unsupervised word
  sense disambiguation.
\newblock {\em IEEE transactions on pattern analysis and machine intelligence}
  32(4):678--692.

\bibitem[\protect\citeauthoryear{Nguyen \bgroup et al\mbox.\egroup
  }{2017}]{nguyen2017robust}
Nguyen, D.~T.; Al~Mannai, K.~A.; Joty, S.; Sajjad, H.; Imran, M.; and Mitra, P.
\newblock 2017.
\newblock Robust classification of crisis-related data on social networks using
  convolutional neural networks.
\newblock In {\em Eleventh International AAAI Conference on Web and Social
  Media}.

\bibitem[\protect\citeauthoryear{Pennington, Socher, and
  Manning}{2014}]{pennington2014glove}
Pennington, J.; Socher, R.; and Manning, C.
\newblock 2014.
\newblock Glove: Global vectors for word representation.
\newblock In {\em Proceedings of the 2014 conference on empirical methods in
  natural language processing (EMNLP)},  1532--1543.

\bibitem[\protect\citeauthoryear{Popescu, Pennacchiotti, and
  Paranjpe}{2011}]{popescu2011extracting}
Popescu, A.-M.; Pennacchiotti, M.; and Paranjpe, D.
\newblock 2011.
\newblock Extracting events and event descriptions from twitter.
\newblock In {\em Proceedings of the 20th international conference companion on
  World wide web},  105--106.
\newblock ACM.

\bibitem[\protect\citeauthoryear{Sakaki, Okazaki, and
  Matsuo}{2010}]{sakaki2010earthquake}
Sakaki, T.; Okazaki, M.; and Matsuo, Y.
\newblock 2010.
\newblock Earthquake shakes twitter users: real-time event detection by social
  sensors.
\newblock In {\em Proceedings of the 19th international conference on World
  wide web},  851--860.
\newblock ACM.

\bibitem[\protect\citeauthoryear{Sankaranarayanan \bgroup et al\mbox.\egroup
  }{2009}]{sankaranarayanan2009twitterstand}
Sankaranarayanan, J.; Samet, H.; Teitler, B.~E.; Lieberman, M.~D.; and
  Sperling, J.
\newblock 2009.
\newblock Twitterstand: news in tweets.
\newblock In {\em Proceedings of the 17th acm sigspatial international
  conference on advances in geographic information systems},  42--51.
\newblock ACM.

\bibitem[\protect\citeauthoryear{Srivastava \bgroup et al\mbox.\egroup
  }{2014}]{srivastava2014dropout}
Srivastava, N.; Hinton, G.; Krizhevsky, A.; Sutskever, I.; and Salakhutdinov,
  R.
\newblock 2014.
\newblock Dropout: a simple way to prevent neural networks from overfitting.
\newblock {\em The Journal of Machine Learning Research} 15(1):1929--1958.

\bibitem[\protect\citeauthoryear{Twitter}{2019}]{GNIP}
Twitter, I.
\newblock 2019.
\newblock Gnip api.

\bibitem[\protect\citeauthoryear{Xie, Kelley, and
  Szymanski}{2013}]{xie2013overlapping}
Xie, J.; Kelley, S.; and Szymanski, B.~K.
\newblock 2013.
\newblock Overlapping community detection in networks: The state-of-the-art and
  comparative study.
\newblock {\em Acm computing surveys (csur)} 45(4):43.

\bibitem[\protect\citeauthoryear{Xie, Szymanski, and Liu}{2011}]{xie2011slpa}
Xie, J.; Szymanski, B.~K.; and Liu, X.
\newblock 2011.
\newblock Slpa: Uncovering overlapping communities in social networks via a
  speaker-listener interaction dynamic process.
\newblock In {\em Data Mining Workshops (ICDMW), 2011 IEEE 11th International
  Conference on},  344--349.
\newblock IEEE.

\bibitem[\protect\citeauthoryear{Yarowsky}{1995}]{yarowsky1995unsupervised}
Yarowsky, D.
\newblock 1995.
\newblock Unsupervised word sense disambiguation rivaling supervised methods.
\newblock In {\em 33rd annual meeting of the association for computational
  linguistics}.

\bibitem[\protect\citeauthoryear{Zhang \bgroup et al\mbox.\egroup
  }{2019}]{zhang2019social}
Zhang, C.; Fan, C.; Yao, W.; Hu, X.; and Mostafavi, A.
\newblock 2019.
\newblock Social media for intelligent public information and warning in
  disasters: An interdisciplinary review.
\newblock {\em International Journal of Information Management} 49:190--207.

\bibitem[\protect\citeauthoryear{Zhou, Chen, and
  He}{2015}]{zhou2015unsupervised}
Zhou, D.; Chen, L.; and He, Y.
\newblock 2015.
\newblock An unsupervised framework of exploring events on twitter: Filtering,
  extraction and categorization.
\newblock In {\em AAAI},  2468--2475.

\end{thebibliography}
\bibliographystyle{aaai}

\newpage

\section{Supplemental Material}
\label{sec:supplemental}
Here is the full list of keywords used for each event category (Section {\it Event Categories and Event Keywords}). 
Various word forms of the keywords are also considered, e.g., ``evacuates, evacuated, evacuating'' are also considered for the keyword ``evacuate''.

\noindent
1) Preventative measure (PRE): evacuate, evacuation, evacuee, shelter, refugee

\noindent
2) Help and rescue (RES): rescue, boat, help, donate, guard

\noindent
3) Casualty (CAS): die, dead, drown, injure, hurt

\noindent
4) Housing (HOU): house, home, room, apt, apartment

\noindent
5) Utilities and Supplies (UTI): power, electricity, gas, store, food, supply

\noindent
6) Transportation (TRA): airplane, plane, flight, airport, ``RoadTypes'' (highway, freeway, road, avenue, ave, dr, rd, st, hwy, fwy, blvd)

\noindent
7) Flood control infrastructures (FCI): reservoir, bayou, canal, dam, levee

\noindent
8) Business Work School (BWS): office, school, closed, open, work

\noindent
9) Built-environment hazards (HAZ): fire, explosion, collapse, poison, electrocute

\end{document}